\documentclass[runningheads]{llncs}
\usepackage{graphicx}
\usepackage{cite}
\usepackage{amsmath,amssymb,amsfonts}
\begin{document}
\title{Kernel Inversed Pyramidal Resizing Network for Efficient Pavement Distress Recognition}
\author{Rong Qin\inst{1}\orcidID{0000-0003-4949-2168} \and
Luwen Huangfu\inst{2,3}\orcidID{0000-0003-3926-7901} \and
Devon Hood\inst{4}\orcidID{0000-0002-3445-0603} \and
James Ma\inst{5}\orcidID{0000-0001-7456-697X} \and
Sheng Huang\inst{1,6,*}\orcidID{0000-0001-5610-0826}}
%


%
\authorrunning{Rong Qin et al.}

\institute{School of Big Data \& Software Engineering, Chongqing University, Chongqing, 400044, China \and
Fowler College of Business, San Diego State University, San Diego, California, 92182, USA \and
Center for Human Dynamics in the Mobile Age, San Diego State University, San Diego, California, 92182, USA \and
College of Science, San Diego State University, San Diego, California, 92182, USA \and
Department of Business Analysis, University of Colorado at Colorado Springs, Colorado, 80918, USA \and
Ministry of Education Key Laboratory of Dependable Service Computing in Cyber Physical Society, Chongqing University, Chongqing, 400044, China\\
\email{\{qinrongzxxlxy,huangsheng\}@cqu.edu.cn,~\{lhuangfu,dhood1037\}@sdsu.edu,~jma@uccs.edu}\\
* indicates corresponding author.}
\maketitle              % typeset the header of the contribution
\begin{abstract}
Pavement Distress Recognition (PDR) is an important step in pavement inspection and can be powered by image-based automation to expedite the process and reduce labor costs. Pavement images are often in high-resolution with a low ratio of distressed to non-distressed areas. Advanced approaches leverage these properties via dividing images into patches and explore discriminative features in the scale space. However, these approaches usually suffer from information loss during image resizing and low efficiency due to complex learning frameworks. In this paper, we propose a novel and efficient method for PDR. A light network named the Kernel Inversed Pyramidal Resizing Network (KIPRN) is introduced for image resizing, and can be flexibly plugged into the image classification network as a pre-network to exploit resolution and scale information. In KIPRN, pyramidal convolution and kernel inversed convolution are specifically designed to mine discriminative information across different feature granularities and scales. The mined information is passed along to the resized images to yield an informative image pyramid to assist the image classification network for PDR. We applied our method to three well-known Convolutional Neural Networks (CNNs), and conducted an evaluation on a large-scale pavement image dataset named CQU-BPDD. Extensive results demonstrate that KIPRN can generally improve the pavement distress recognition of these CNN models and show that the simple combination of KIPRN and EfficientNet-B3 significantly outperforms the state-of-the-art patch-based method  in both performance and efficiency.

\end{abstract}
\section{Introduction}
Pavement distress is one of the largest threats to modern road networks and, as such, Pavement Distress Recognition (PDR) is an important aspect in maintaining logistics infrastructure. Traditionally, pavement distress recognition is done manually by professionals, and this requires a large overhead of labor and extensive domain knowledge~\cite{benedetto2014fdtd}. Given the complex and vast network of roadways, it is almost impossible to accomplish the pavement distress inspection manually. Therefore, automating the pavement distress recognition task is essential.

In recent decades, many methods have been proposed to address this issue through the use of computer vision. Conventional approaches often utilize rudimentary image analysis, hand-crafted features, and traditional classifiers~\cite{salman2013pavement, wang2010pavement, chou1994pavement, li2008novel, nejad2011expert}. For example, Salman et al.~\cite{salman2013pavement} proposed a crack detection method based on the Gabor filter, and Li et al.~\cite{li2008novel} developed a neighboring difference histogram method to detect conventional, human visual, pavement disease. The main problem of these approaches is that the optimizations of feature extraction and image classification steps are separated or even omitted from any learning process. Inspired by the remarkable success of deep learning, numerous researchers have employed such models to solve this problem~\cite{gopalakrishnan2017deep, fan2019road, naddaf2020efficient, li2020automatic, tang2020iteratively, bhagvati1994gaussian}. For example, Gopalakrishnan et al.~\cite{gopalakrishnan2017deep} employed a Deep Convolutional Neural Network (DCNN) trained on the ImageNet database, and transferred that learning to automatically detect cracks in Hot-Mix Asphalt (HMA) and Portland Cement Concrete (PCC) surfaced pavement images. Fan et al.~\cite{fan2019road} proposed a novel road crack detection algorithm that is based on deep learning and adaptive image segmentation. However, these approaches often overlook many key characteristics of pavement images, such as the high image resolution and the low ratio of distressed to non-distressed areas.

\begin{figure}[t]
\centering
\includegraphics[scale=0.33]{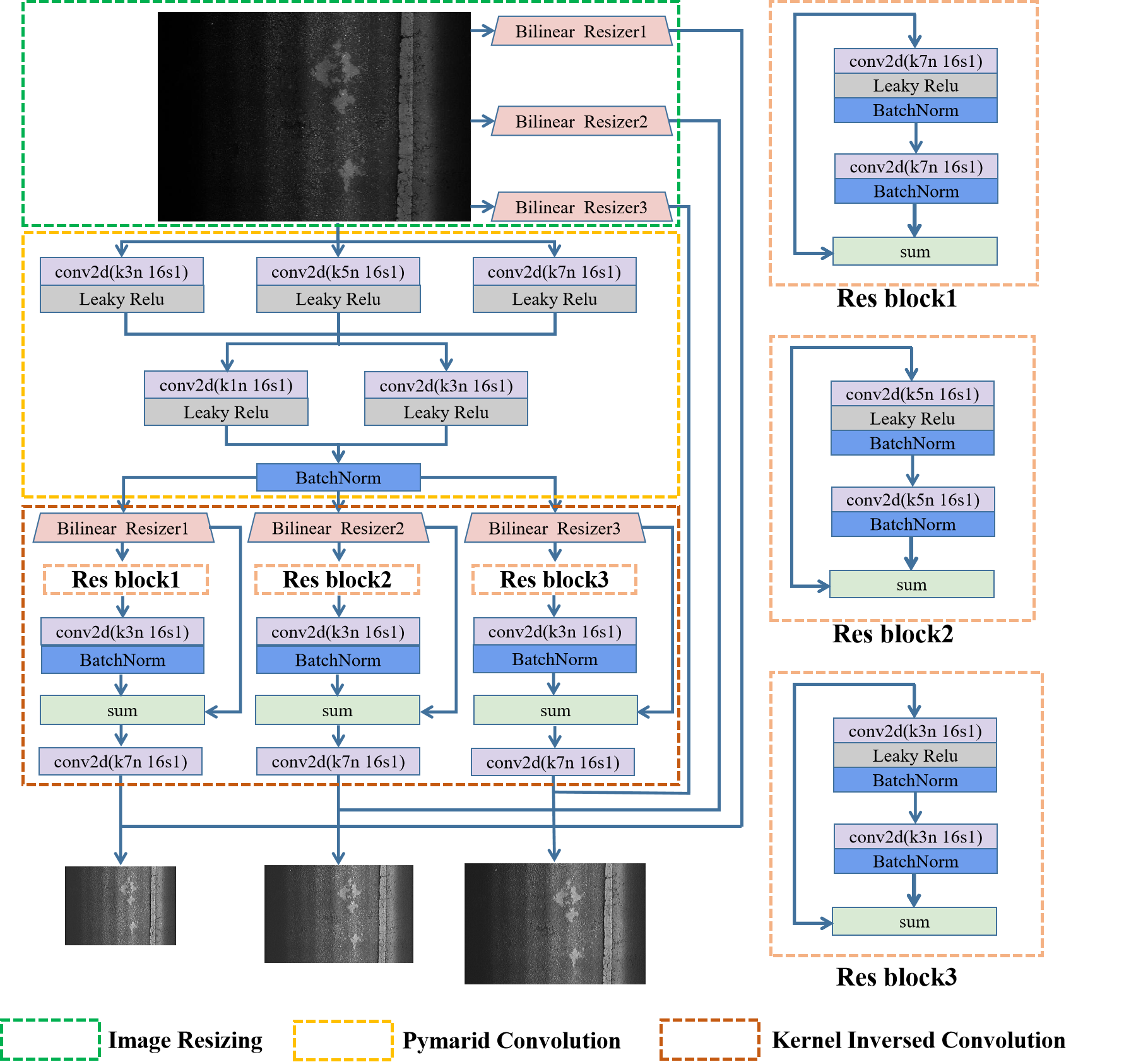}
\vspace{-0cm}
\caption{The network architecture of Kernel Inversed Pyramidal Resizing Network (KIPRN). %KIPRN consists of three modules: Image Resizing (IR), Pyramidal Convolution (PC), and Kernel Inversed Convolution (KIC). PC leverages the pyramidal convolutions to extract the information in different feature granularity, and KIC mines the discriminative information of image in different scale with different branches. At last, these learned information will be compensated to the images resized by IR to yield the final image pyramid.
}
\label{frame work}
\vspace{-0.7cm}
\end{figure}

Huang~et al~\cite{huang2021weakly} presented a Weakly Supervised Patch Label Inference Network with Image Pyramid (WSPLIN-IP) to solve those problems. The WSPLIN-IP exploited high resolution information and scale space of images by dividing an image pyramid into patches for label inference via weakly supervised learning and achieved promising performances in comparison with other state-of-the-art approaches. However, its patch collection strategy and the complex patch label inference processes can lead to low efficiency in practical applications. Moreover, as the mainstream PDR approach, the CNN-based methods often need to resize images to a uniform size for the CNNs, where traditional resizing algorithms, such as bilinear interpolation, are employed. As a result, image resizing is completely independent of model optimization, which inevitably leads to the loss of some discriminative information. A few existing related studies, such as~\cite{recasens2018learning}, are often difficult to apply to pavement images due to the high sensitivity of pavement diseases to deformation and the need for multi-scale input.

Inspired by the idea of Resizing Network~\cite{talebi2021learning}, we elaborate a light image resizing network named the Kernel Inversed Pyramidal Resizing Network (KIPRN) to address these issues. The KIPRN can be integrated into any deep learning-based model as a self-contained supplemental module and be optimized  with the model as one whole integration, and it learns to retain the information, and compensates for the information loss caused by the image resizing based on the bilinear interpolation.
As shown in Figure~\ref{frame work}, KIPRN employs pyramidal convolutions~\cite{duta2020pyramidal} to extract information from the original pavement images with different granularity, and then produces a three-layer image pyramid for each input image with our designed kernel inversed ResBlock. Pyramidal convolutions enable the mining of more resolution information with different sizes of convolutions, while the kernel inversed ResBlock can better mine the scale information by enlarging the differences between relative receptive fields in different resizing branches. Finally, the produced image pyramid will be input into the subsequent deep learning-based PDR model to exploit the scale space again without greatly increasing computational burdens.

We evaluated the KIPRN on a large-scale pavement image dataset named CQU-BPDD~\cite{tang2020iteratively}. Extensive results show that our method generally boosts many deep learning-based PDR models. Moreover, our enhanced EfficientNet-B3 not only achieved state-of-the-art performances, but also obtained prominent advantages in efficiency compared with the WSPLIN-IP, which is a recent state-of-the-art PDR approach that also considers the resolution and scale information and utilizes the EfficientNet-B3 as its backbone network. The main contributions of our work can be summarized as follows:

$\bullet$  We propose a novel resizing network named KIPRN that can boost any deep learning-based PDR approach by exploiting the resolution and scale information of images. Moreover, the KIPRN will not require significant computational cost. To the best of our knowledge, our work is the first attempt to use a deep learning model to study image resizing in pavement distress analysis.
%We propose an end-to-end lightweight convolution neural network named KIPRN. It significantly improves the effect of classifier on pavement distress recognition with minimal loss of inferring speed and has great generalization in pavement distress recognition.

$\bullet$ We propose kernel inversed ResBlocks, which applies the smaller convolutional kernels to the larger feature maps, while applying the larger convolutional kernels to the smaller feature maps in a size-inversion way. It implicitly enlarges the scale space and, thereby, enhances the scale information exploitation.
 %It further enhances the difference of the relative receptive field and achieve the competitive performances in comparison with other designs

$\bullet$ Extensive results demonstrate that the KIPRN can generally improve deep learning-based PDR approaches and achieve a state-of-the-art performance without greatly increasing computational burdens.
%We empirically study several classical computer vision models, WSPLIN-IP~\cite{huang2021weakly} and KIPRN on a large-scale dataset. Extensive results show that KIPRN significantly outperforms them and can improve a variety of computer vision models.
%
\section{Methodology}
\subsection{Problem Formulation and Overview}
Pavement distress recognition is an image classification task to classify the images of damaged pavements into different distress categories. Let $X = \{ x_i\}_{i=1}^{n}$ and $Y = \{ y_i\}_{i=1}^{n}$ be the pavement images and their corresponding labels respectively. $y_i$ is a $C$-dimensional one-hot vector, where $C$ is the number of pavement distress categories. $y_{ij}$ represents the $j$-th element of $y_i$. If the $j$-th element is the only non-zero element, it indicates that the corresponding pavement image has the $j$-th type of pavement distress. The goal of pavement distress recognition is to train a PDR model to recognize pavement distress in a given pavement image.

In deep learning-based PDR, the high-resolution images are often resized into a fixed size to meet the input or efficiency requirement of these models. Moreover, some studies also show that exploiting the scale information of pavement images can benefit pavement distress recognition. Thus, image resizing is an inevitable process in pavement distress recognition based on deep learning. However, the conventional linear interpolation-based image resizing process is independent of the optimization of the pavement distress model, and, thereby, often causes the loss of discriminative information.

To address this issue, we propose an end-to-end network named KIPRN for training to resize the pavement images and, thereby, aiding the PDR model. KIPRN can be plugged into any deep learning-based image classification models as a pre-network and optimized with these models. The pavement distress recognition process based on the KIPRN can be represented as follows,
\begin{equation}\label{}
\hat{y} \leftarrow {\rm softmax}(P_\phi(\Gamma_\theta(x))),
\end{equation}
where $\Gamma(\cdot)$ and $P(\cdot)$ are the mapping functions of the KIPRN and the PDR model respectively, while $\phi$ and $\theta$ are their corresponding parameters. As shown in Figure~\ref{frame work}, the KIPRN consists of three modules, namely Image Resizing (IR), Pyramidal Convolution (PC) and Kernel Inversed Convolution (KIC). On the one hand, the original pavement images will be resized into different sizes through IR. On the other hand, PC extracts the features of original pavement images from different granularities, and then KIC mines the scale information of the extracted features with different branches via inversed kernels. The mined information will be compensated into the resized images to yield a three-layer image pyramid for each pavement image. Finally, the produced image pyramid will be input into the subsequent deep learning-based PDR model to accomplish the label inference.

\subsection{Image Resizing}
The Image Resizing (IR) module is used to resize the original pavement image into $m$ different sizes. Here, we set $m=3$. This process can be represented as follows,
\begin{equation}\label{}
\mathcal{I} = H(x),
\end{equation}
where $H(\cdot)$ is any chosen traditional image resizing algorithm, and $\mathcal{I}=\{ I_j\}_{j=1}^{m}$ is the collection of resized images. $I_j$ is the $j$-th resized pavement image. We followed~\cite{talebi2021learning} and adopted bilinear interpolation as the image resizing algorithm. The KIPRN will compensate the subsequently learned discriminative and scale information into $\mathcal{I}$ to generate a pavement image pyramid that is more conducive for a deep learning-based PDR model to correctly recognize pavement distress.

%These resized images $\mathcal{I}=\{ I_j\}_{j=1}^{m}$ will be used for constructing the image pyramid for assisting the pavement distress recognition model to further exploit the scale information.

\subsection{Pyramidal Convolution}
The next step of the KIPRN is to leverage a Pyramidal Convolution (PC) module to mine and preserve the relevant information of the original images from different feature granularities. The golden dash-line rectangle in Figure~\ref{frame work} shows the details of the PC module, which is a two-layers of pyramidal convolution~\cite{duta2020pyramidal}. The first layer adopts three convolution kernels, whose sizes are $3\times3$, $5\times5$, and $7\times7$ respectively, while two convolution kernels of the second layer are $1\times1$ and $3\times3$ respectively.  Let $Q(\cdot)$ be the mapping function of the PC module where $\eta$ is its corresponding parameter. The pyramidal feature map can be generated as follows,
\begin{equation}\label{}
f = Q_\eta(x),
\end{equation}
which sufficiently encodes the detailed features of pavement images under different granularities.
%where $Q(\cdot)$ is the mapping functions of PCM, $\eta$ is its learnable parameters and $f$ represents the extracted different features of the original pavement image. Intuitively speaking, pyramidal convolution uses convolution kernels of different sizes to extract multiple sets of feature maps, where smaller convolution kernels corresponds to mappings of more channels and vice versa. The final feature maps are then spliced together in order to combine the different features highlighted by each kernel. The following may represent the mapping function of the pyramidal convolution,
%\begin{equation}\label{}
%pyconv(\mathcal{F}) = \left\{conv_1(\mathcal{F}),\cdots,conv_k(\mathcal{F})\right\},
%\end{equation}
%where $\mathcal{F}$ is the input feature map, $pyconv$ is the mapping functions of pyramidal convolution and $ \left\{conv_1,\cdots,conv_k\right\}$ are the mapping functions of convolution kernels with different sizes.

\subsection{Kernel Inversed Convolution}
Once we obtain the feature map $f$, a Kernel Inversed Convolutional (KIC) module is designed to better exploit the scale information of pavement images by enlarging the differences in the receptive fields in different convolution branches, as shown in the red dash-line rectangle in Figure~\ref{frame work}. In KIC, the feature map $f$ is resized into $m$ different sizes, which are identical to the ones of those resized images in the image resizing module,
\begin{equation}\label{}
\hat{F}=H(f)=\{\hat{f}_j\}^m_{j=1},
\end{equation}
where $\hat{F}=\{\hat{f}_j\}^m_{j=1}$ is the collection of the resized feature maps, and $\hat{f}_j$ is the resized feature map corresponding to the resized image $I_j$. Thereafter, these resized feature maps $\hat{F}$ are fed into different convolution branches to produce the information compensations for different resized images, and yield the final image pyramid. The whole image pyramid generation process can be represented as follows,
\begin{equation}\label{}
S = R_\zeta(\hat{F})+\mathcal{I}=\{s_j\}_{j=1}^m,
\end{equation}
where $S=\{s_j\}_{j=1}^m$ is the generated image pyramid, $R(\cdot)$ is the mapping function of the Kernel Inversed Convolution (KIC) module, and $\zeta$ is its learned parameters.

To better retain the discriminative information of pavement images across different scales, we adopted an idea from ~\cite{chen2018big}, and elaborate a series of kernel inversed ResBlocks for feature learning. In these ResBlocks, as shown in the orange dash-line rectangles in Figure~\ref{frame work}, the smaller convolution kernel is applied to the larger feature map, which enables the mining of the local detailed information of images, while the larger convolution kernel is applied to the smaller feature map, which enables the capturing of the global structural features of images. In other words, the kernel inversed convolution enlarges the perception range in the scale space, and thereby preserves more conducive information for the solution of the subsequent task. In this manner, the whole KIPRN process is a composition of these modules, $\Gamma=H\circ Q \circ R$, where $\theta=\{\eta,\zeta\}$.

%In order to make the generated image pyramid more conducive to pavement distress recognition, similar to work~\cite{chen2018big}, for features of different sizes, we design kernel inversed Resblocks in KICM to better extract scale information. The difference however, since the convolution layer after Resblock is the same, is that KICM does not use different channel numbers for different Resblocks, but instead use convolution kernels of different sizes. For a convolution kernel, the larger the feature, the smaller the relative receptive field will be. As such, the purpose of amplifying features is similar to using a smaller convolution kernel. Therefore, in the kernel inversed Resblocks, for a larger feature, we choose a Resblock with a smaller convolution kernel to further enhance the differences between the relative receptive fields in different resizing branches which can better mine the scale information. The addition of our kernel inversed Resblock design shows impressive improved results in our experimental study as described later in this paper.

\subsection{Multi-Scale Pavement Distress Recognition}
In the final step, the generated image pyramid was input into an image classification network $P_\phi(\cdot)$ to predict the distress label of a given pavement image as follows,
\begin{equation}\label{}
\hat{y} \leftarrow {\rm softmax}(\sum_{j=1}^m{P(s_j)})={\rm softmax}({P(R(H(Q(x)))+H(x))}).
\end{equation}
The KIPRN was optimized with the image classification network together in an end-to-end manner. Let $\mathcal{L}(\cdot,\cdot)$ be the loss function of the image classification network. The optimal parameters of the KIPRN and the image classification network can be obtained by solving the following programming problem,
\begin{equation}\label{}
\{\hat{\theta},\hat{\phi}\} \leftarrow  \arg\underset{\theta,\phi}\min \sum_{i=1}^{n}\mathcal{L}(y_i,\hat{y}_i).
\end{equation}
In this study, we chose three well known Convolutional Neural Networks (CNNs), namely EfficientNet-B3, ResNet-50 and Inception-v1, as the image classification networks to validate the effectiveness of the KIPRN.
\section{Experiment}
\subsection{Dataset and Setup}
A large-scale pavement dataset named CQU-BPDD~\cite{tang2020iteratively} was employed for evaluation. It consists of 43,861 typical pavement images and 16,795 diseased pavement images and includes seven different types of distresses, namely alligator crack, crack pouring, longitudinal crack, massive crack, transverse crack, raveling, and repair. Since pavement distress recognition is a follow-up task of pavement distress detection, we only used the diseased pavement images for training and testing. For a fair comparison, we followed the data split strategy in~\cite{huang2021weakly}, so that 5,140 images were selected for training while the remaining ones were used for testing.

As shown in Table~\ref{result}, we chose three traditional shallow learning methods, five well-known CNNs, two classical vision transformers and WSPLIN-IP as baselines. %They are Histogram of Oriented Gradient (HOG)~\cite{dalal2005histograms}, Support Vector Machine (SVM)~\cite{cortes1995support} , Random Forest (RF)~\cite{breiman2001random},  VGG-16~\cite{simonyan2014very}, ResNet-50~\cite{he2016deep}, Inception-v1~\cite{szegedy2015going}, Inception-v3~\cite{szegedy2016rethinking}, EfficientNet-B3~\cite{tan2019efficientnet}, ViT-S/16~\cite{dosovitskiy2020image} ViT-B/16~\cite{dosovitskiy2020image} and WSPLIN-IP~\cite{huang2021weakly}.
For all deep learning models, we adopted AdamW~\cite{loshchilov2018fixing} as the optimizer and the learning rate was set to 0.0001. The input images were resized by the KIPRN into three resolutions, $300\times300$, $400\times400$, and, $500\times500$, to yield the image pyramid. Recognition accuracy is used as the performance metric following~\cite{huang2021weakly}.

\begin{table}[h]

	\centering
	\vspace{-0.4cm}
   % \fontsize{12}{12}\selectfont
	\caption{The recognition accuracies of different methods on CQU-BPDD.}
	\vspace{-0.3cm}
%  	\resizebox{\columnwidth}{!}{
		\begin{tabular}{c c c}
			\hline
			Method & Type & Accuracy \\
			\hline
            RGB + RF~\cite{breiman2001random} & Single-scale & 0.305 \\
			
			HOG~\cite{dalal2005histograms} + SVM~\cite{cortes1995support} & Single-scale & 0.318 \\
			
			VGG-16~\cite{simonyan2014very} & Single-scale & 0.562  \\

			Inception-v3~\cite{szegedy2016rethinking} & Single-scale & 0.716  \\

			ViT-S/16~\cite{dosovitskiy2020image} & Single-scale & 0.750  \\
			
			ViT-B/16~\cite{dosovitskiy2020image} & Single-scale & 0.753   \\
			
			WSPLIN-IP\thefootnote{}~\cite{huang2021weakly} & Multi-scale & 0.837 \\
            WSPLIN-IP~\cite{huang2021weakly} & Multi-scale & 0.850 \\
            \hline
			ResNet-50~\cite{he2016deep} & Single-scale & 0.712 \\
			ResNet-50 & Multi-scale &  0.786\\
			\textbf{Ours} + ResNet-50 & Multi-scale & 0.827  \\
			\hline			
			Inception-v1~\cite{szegedy2015going} & Single-scale & 0.726  \\
			Inception-v1 & Multi-scale &  0.781 \\
			\textbf{Ours} + Inception-v1 & Multi-scale & 0.803  \\
			
			\hline
			
            EfficientNet-B3~\cite{tan2019efficientnet} & Single-scale & 0.786  \\
            EfficientNet-B3 & Multi-scale & 0.830  \\
			\textbf{Ours} + EfficientNet-B3 & Multi-scale & \textbf{0.861}\\
			%\hdashline
			%Gains & Multi-size & {\color{red}+}{\color{red}0.031}\\
			
			%\hdashline
			%Gains & Multi-size & {\color{red}+}{\color{red}0.022}\\
			\hline
		\end{tabular}
%  	}
	\label{result}
	\vspace{-0.7cm}
\end{table}

\begin{figure}[h]
%\begin{tabular}{cc}
%\vspace{-2cm}
\begin{minipage}{0.48\linewidth}
  \centerline{\includegraphics[width=3.8cm,height=2.2cm]{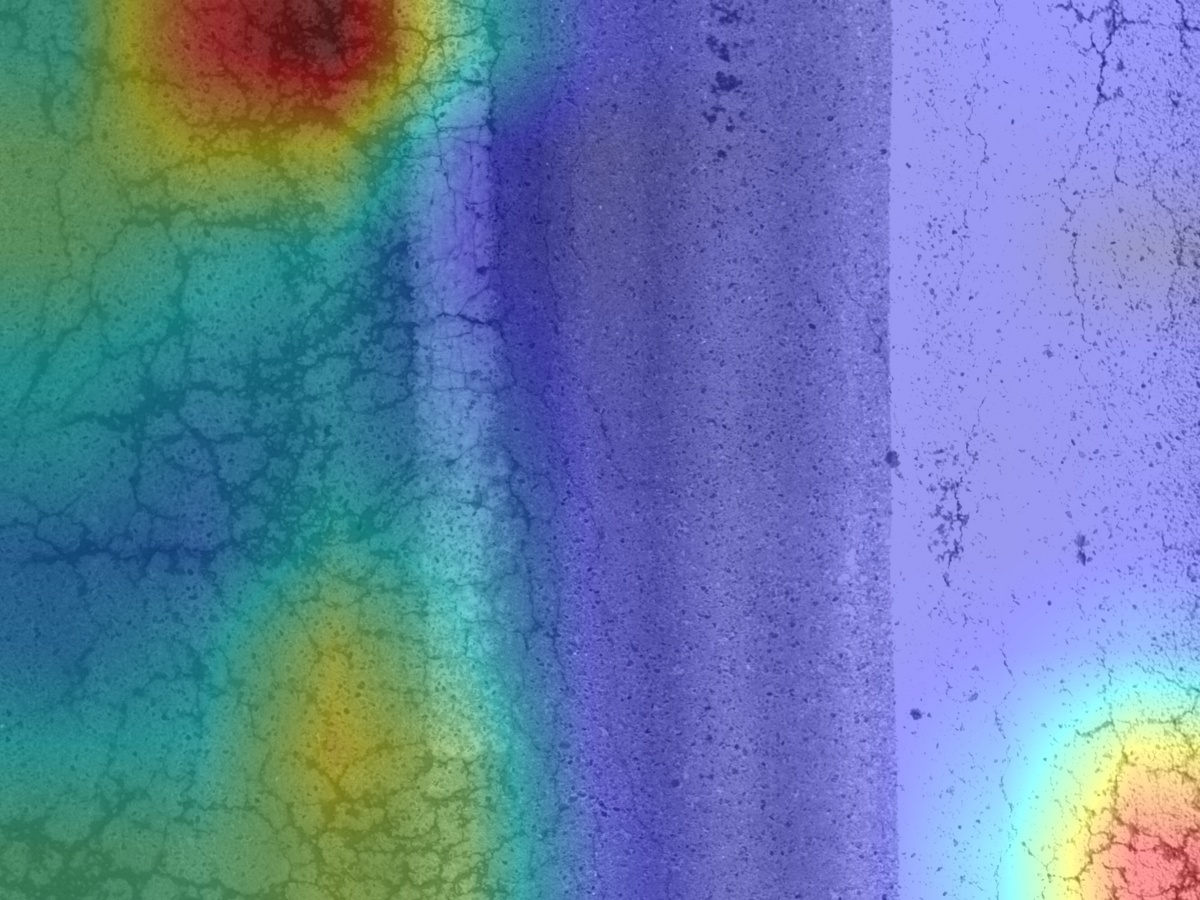}}
  \centerline{(a) EfficientNet-B3}
\end{minipage}
\hfill
\begin{minipage}{.48\linewidth}
  \centerline{\includegraphics[width=3.8cm,height=2.2cm]{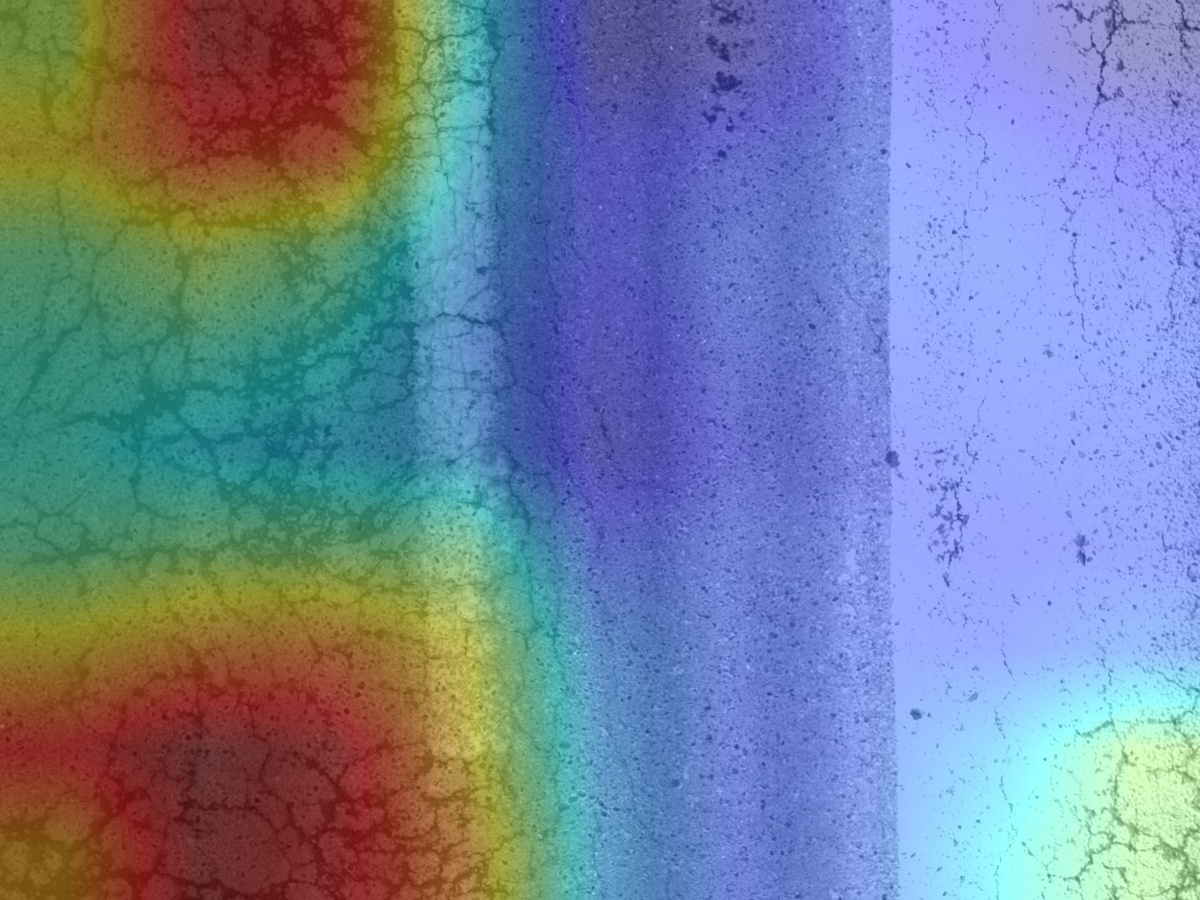}}
  \centerline{(b) EfficientNet-B3 + KIPRN}
\end{minipage}
\vfill
\begin{minipage}{0.48\linewidth}
  \centerline{\includegraphics[width=3.8cm,height=2.2cm]{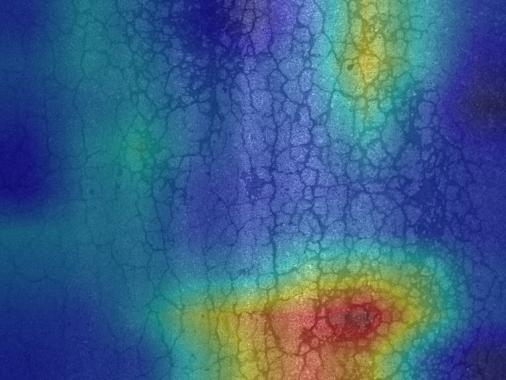}}
  \centerline{(c) EfficientNet-B3}
\end{minipage}
\hfill
\begin{minipage}{0.48\linewidth}
  \centerline{\includegraphics[width=3.8cm,height=2.2cm]{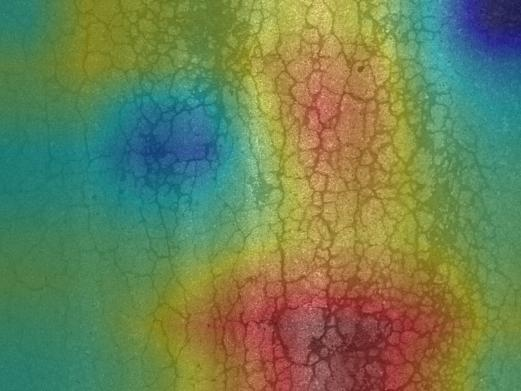}}
  \centerline{(d) EfficientNet-B3 + KIPRN}
\end{minipage}
%\end{tabular}
\vspace{-0.3cm}
\caption{Comparison of Class Activation Mapping (CAM) between EfficientNet-B3 and EfficientNet-B3 + KIPRN.}
\label{fig:cam}
\vspace{-0.3cm}
\end{figure}

\begin{figure}[h]
\centering
%\vspace{-0.2cm}
\includegraphics[scale=0.35]{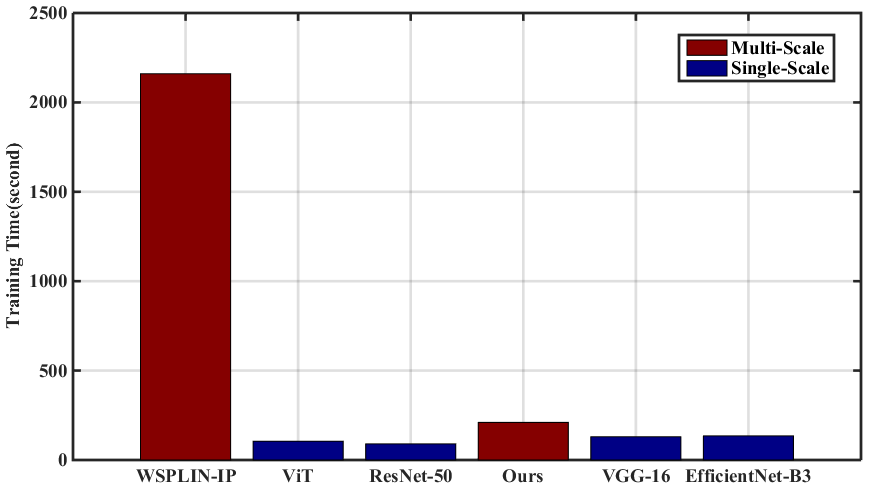}
\vspace{-0.3cm}
\caption{The training time of models per epoch on the CQU-BPDD dataset.}
\label{time}
\vspace{-0.5cm}
\end{figure}
%\vspace{-0.5cm}
\footnotetext{The reproduced results with the codes provided by the original authors under our experimental settings.}
%\vspace{-0cm}
\begin{table}[h]
	\centering
	%\fontsize{15}{14.5}\selectfont
%	\vspace{-0.2cm}
	\caption{Different settings of ResBlocks and pyramidal convolution in the KPRIN deployed with EfficientNet-B3. The i × i indicates the size of convolution kernel in Resblock. Forward is the strategy that applies a small kernel to the small feature map. Inversed is the inversed strategy of Forward, which is the strategy adopted in our method. Pyconv is pyramidal convolution. First and Last puts Pyconv on the first or the last two convolution layers.}
	\vspace{-0.3cm}
	%\resizebox{\columnwidth}{!}{
% \normalsize
		\begin{tabular}{c c c}
			\hline
			Moudule & Setting/Design & Accuracy\\
			\hline
            Resblocks & 3×3 & 0.844\\

            Resblocks & 5×5 & 0.852\\

            Resblocks & 7×7 & 0.847\\

            Resblocks & Foward & 0.857\\
			
            Resblocks & Inversed & \textbf{0.861}\\
            \hline

            Pyconv & All & 0.828\\

            Pyconv & None & 0.855\\

            Pyconv & Resblock & 0.845\\

            Pyconv & Last & 0.830\\

            Pyconv & First & \textbf{0.861}\\

			\hline
		\end{tabular}
\vspace{-0.5cm}
	%}
	\label{result2}
	
\end{table}
\subsection{Pavement Disease Recognition}
Table~\ref{result} tabulates the pavement distress recognition accuracies of different methods on the CQU-BPDD dataset. The results show that the KIRPN significantly boosts all three chosen CNN-based image classification models. The KIPRN improves the the accuracy of EfficientNet-B3, ResNet-50 and Inception-v1 by $7.5\%$, $11.5\%$ and $7.7\%$ respectively. Moreover, EfficientNet-B3 enhanced by the KIPRN achieved the best performance among all the methods. As shown in Figure~\ref{fig:cam}, the Class Activation Mapping (CAM) of the EfficientNet-B3 enhanced by the KIPRN is better than that of the original EfficientNet-B3. Another interesting observation regarding the multi-scale approaches is they often perform better than their single-scale versions, even though they simply use bilinear interpolation to resize images. The second best performed method, WSPLIN-IP, another multi-scale approach, adopted EfficientNet-B3 as its backbone networks. However, EfficientNet-B3 enhanced by KIPRN achieved 2.4\% accuracy gains over WSIPLIN-IP under the same experimental settings. Overall, the multi-scale approaches based on the KIPRN consistently outperformed the versions based on the bilinear interpolation. These results clearly reveal two facts, namely 1$)$ exploiting scale space of images can improve pavement distress recognition, and 2$)$ the KIPRN enables learning conductive information for distress recognition during image resizing.

\subsection{Ablation Study}

Table~\ref{result2} tabulates the impacts of different convolution settings on the performance of the KIPRN deployed on EfficientNet-B3. These experiments were conducted on KIPRN deployed on EfficientNet-B3. The results show that the kernel inversed strategy outperforms all the other ResBlock settings. This verifies that our strategy can capture more information in the scale space by enlarging the differences in relative receptive fields. Another interesting observation is that putting pyramidal convolution on the first two layers led to the best performances. We attribute this to the fact that pyramidal convolutions are more capable of capturing low-level visual details than learning the abstract semantic features that the last layers are designed for.

Figure~\ref{time} reports the training times of different models per epoch. The observations show that EfficientNet-B3 enhanced by the KIPRN (ours) has similar training efficiency compared to the single-scale EfficientNet-B3, while enjoying the 10X training speeds over WISPLIN-IP, which is also a multi-scale approach.

\section{Conclusion}
In this study, we proposed an end-to-end resizing network named the KIPRN, which can boost any deep learning-based PDR approach by assisting it to better exploit the resolution and scale information of images. The KIPRN consists of IR, PC and KIC. The  KIPRN can be integrated into any deep learning-based PDR model in a plug-and-play way and optimizes together with a deep learning-based PDR model in an end-to-end manner. Extensive results show that our method generally boosts many deep learning-based PDR models. In our future work, we will attempt to compensate the features to the original pavement image in a better way than simple addition.

\section{Acknowledgments}
This study was supported by the San Diego State University 2021 Emergency Spring Research, Scholarship, and Creative Activities (RSCA) Funding distributed by the Division of Research and Innovation, as well as XSEDE EMPOWER Program under National Science Foundation grant number ACI-1548562.

\bibliographystyle{splncs04.bst}

\bibliography{ref}
\end{document}